\documentclass{article}

\PassOptionsToPackage{numbers, compress}{natbib}
\PassOptionsToPackage{dvipsnames}{xcolor}

\usepackage[preprint]{neurips_2023}

\usepackage[utf8]{inputenc} % allow utf-8 input
\usepackage[T1]{fontenc}    % use 8-bit T1 fonts
\usepackage{hyperref}       % hyperlinks
\usepackage{url}            % simple URL typesetting
\usepackage{booktabs}       % professional-quality tables
\usepackage{amsfonts}       % blackboard math symbols
\usepackage{nicefrac}       % compact symbols for 1/2, etc.
\usepackage{microtype}      % microtypography
\usepackage{xcolor}         % colors

\usepackage[bb=dsserif]{mathalpha}
\usepackage{multirow}
\usepackage{multicol}
\usepackage{comment}
\usepackage[normalem]{ulem}
\usepackage{enumitem}
\setlist[itemize]{align=parleft,left=3pt}
\usepackage{graphicx}
\usepackage{subfigure}
\usepackage{amsmath}
\usepackage{amssymb}
\usepackage{mathtools}
\usepackage{amsthm}
\usepackage{wrapfig}
\usepackage{floatrow}
\newcommand{\hl}[1]{\textcolor{TealBlue}{#1}}
\newtheorem{theorem}{Theorem}[section]

\newtheorem{definition}[theorem]{Definition}

\title{On Practical Aspects of Aggregation Defenses \\ against Data Poisoning Attacks}

\author{%
  Wenxiao Wang\\
  Department of Computer Science\\
  University of Maryland\\
  College Park, MD 20742 \\
  \texttt{wwx@umd.edu} \\
  \And
  Soheil Feizi \\
  Department of Computer Science\\
  University of Maryland\\
  College Park, MD 20742 \\
  \texttt{sfeizi@umd.edu} \\
}

\begin{document}

\maketitle

\begin{abstract}
The increasing access to data poses both opportunities and risks in deep learning, as one can manipulate the behaviors of deep learning models with malicious training samples. Such attacks are known as data poisoning. Recent advances in defense strategies against data poisoning have highlighted the effectiveness of aggregation schemes in achieving state-of-the-art results in certified poisoning robustness. However, the practical implications of these approaches remain unclear. Here we focus on Deep Partition Aggregation, a representative aggregation defense, and assess its practical aspects, including efficiency, performance, and robustness. For evaluations, we use ImageNet resized to a resolution of 64 by 64 to enable evaluations at a larger scale than previous ones. Firstly, we demonstrate a simple yet practical approach to scaling base models, which improves the efficiency of training and inference for aggregation defenses. Secondly, we provide empirical evidence supporting the data-to-complexity ratio, i.e. the ratio between the data set size and sample complexity, as a practical estimation of the maximum number of base models that can be deployed while preserving accuracy. Last but not least, we point out how aggregation defenses boost poisoning robustness empirically through the poisoning overfitting phenomenon, which is the key underlying mechanism for the empirical poisoning robustness of aggregations. Overall, our findings provide valuable insights for practical implementations of aggregation defenses to mitigate the threat of data poisoning.
\end{abstract}

\section{Introduction}
\label{sec:intro}

The many great success of deep learning in the fields of vision\cite{ResNet, vit}, natural language\cite{Bert, GPT4}, and speech\cite{SpeechRecog} are only made possible with large amounts of training data. With the amount of training data increasing from thousands \cite{cifar, marcus1993building} to millions or even billions\cite{ImageNet, GPT4, LAION5B}, it is becoming impossible to ensure the reliability of every training sample and therefore the learned models are subject to the threat of data poisoning attacks\cite{WebScalePoison, DatasetSecurity}.

Data poisoning considers an attacker who injects adversarially crafted data into the training set for malicious purposes, including but not restricted to creating targeted misclassification and degrading overall performance. Many defenses have been proposed over the years to mitigate the threat of data poisoning, which can be loosely categorized into defenses through detecting\&filtering poisoned training samples\cite{SteinhardtKL17, DiakonikolasKK019, AnomalyDetect, SpectralSignature, NeuralCleanse, ASSET_detect, SiftOut}, and defenses training directly with poisoned training data\cite{RAB, BackdoorRS, RS_LabelFlipping, DP_defense, GradientShaping, DPA, RobustNN, Bagging, RandomSelection, Adv_Unlearning, CollectiveRobust, FA, ROE, LDC}.
But it is not only the defenses that improve: Many different, potentially adaptive poisoning attacks have been emerging\cite{PoisonSVM, Luis, KSL17, FeatureCollision, ConvexPolytope, Bullseye, Brew, TargetedBackdoor, LC_backdoor, HT_backdoor, freq_trigger, sleeper_agent, Narcissus}, motivating the study of certified defenses.

Recently, the expanding family of aggregation defenses\cite{DPA, Bagging, RandomSelection, CollectiveRobust, FA, ROE, LDC, temporal_robustness} attract much attention for being state-of-the-art in certified robustness against general data poisoning.
However, the practical implications of aggregation defenses remain unclear. In this work, using Deep Partition Aggregation\cite{DPA} as the representative of aggregation defenses, we evaluate aggregations on ImageNet64$\times$64\cite{imagenet64}, with a  larger scale than datasets used by existing evaluations of certified poisoning defenses. We present many techniques and insights regarding practical aspects of aggregation defenses, including efficiency, performance, and robustness.
The contributions include
\begin{itemize}
\setlength\itemsep{0em}
\item demonstrating a simple yet practical approach to scaling base models, which improves the efficiency of training and inference for aggregation defenses;
\item providing empirical evidence supporting the data-to-complexity ratio, i.e. the ratio between the data set size and sample complexity, as a practical estimation of the maximum number of base models that can be deployed while preserving accuracy;
\item uncovering how aggregation defenses boost poisoning robustness empirically through the poisoning overfitting phenomenon, which is the key underlying mechanism for the empirical poisoning robustness of aggregations.
\end{itemize}
\section{Background: Aggregations as Certified Defenses against Data Poisoning}
\label{sec:background}

In this section, we review the designs and the guarantees of Deep Partition Aggregation\cite{DPA} as a representative of aggregations serving as certified defenses against data poisoning attacks.

\textbf{Deep Partition Aggregation\cite{DPA}:} DPA uses the majority votes of classifiers trained from disjoint subsets of training data. 
Let $X$ be the input space, $Y$ be the label space and $D$. 
Given a hyperparameter $k$ as the number of partitions, a deterministic base learner $f$
and a hash function $h: X \times Y \to \mathbb{Z}$ mapping labeled samples to integers, the predictions from DPA is as as follows:
\begin{align*}
    \text{DPA}_{D} (x_0) = \arg\max_{y} \text{DPA}_D(x_0, y) = \arg \max_{y\in \mathcal{Y}} \frac{1}{k} \sum_{i=0}^{k-1}  \mathbb{1}\left[  f_{P_i} (x_0) = y \right],
\end{align*}
where $P_i = \{ (x,y)\in D \mid h(x,y) \equiv i \pmod k \}$ contains all training samples in the $i$-th partition,
$f_{P_i}$ denotes the classifier obtained from the learner $f$ using $P_i$ as training data,
and $\text{DPA}_D(x_0, y)=\frac{1}{k} \sum_{i=0}^{k-1} \mathbb{1}\left[  f _{P_i}(x_0) = y \right]$ denotes the average votes count for class $y$. Ties are broken by returning the smaller class index in $\arg \max$.

\begin{theorem}[Certified Poisoning Robustness of DPA\cite{DPA}]
\label{th:DPA}
Given a training set $D$ and an input $x_0$, let $y_0 = \text{DPA}_{D}(x_0)$, then $\text{DPA}_{D'}(x_0) = y_0$ for any training set $D'$ with
\begin{align}
    |D - D'| \leq \frac{k}{2} \left( \text{DPA}_D(x_0, y_0) - \max_{y\neq y_0} \left( \text{DPA}_D(x_0, y) + \frac{\mathbb{1}\left[y<y_0\right]}{k} \right)\right),
\label{eq:DPA}
\end{align}
where $|D - D'| = |(D\setminus D')\cup (D'\setminus D|$ denotes the symmetric distance between $D$ and $D'$ (i.e. the minimum number of insertions and removals needed to change one set to the other).
\end{theorem}

DPA is a representative of aggregations for certified defenses against data poisoning attacks, and is a perfect example to explain the intuitions behind the certifications: By training each classifier using a smaller amount of samples, each sample can only affect a bounded number of classifiers and therefore adversaries can only impact a bounded number of votes through each poisoned sample; Thus, as in Theorem \ref{th:DPA}, the predictions must remain unchanged unless sufficient samples were poisoned.

Note that aggregation defenses protect not only from triggerless attacks\cite{PoisonSVM, Luis, KSL17, FeatureCollision, ConvexPolytope, Bullseye, Brew}, where there is no modification to test samples, but also from backdoor attacks\cite{TargetedBackdoor, LC_backdoor, HT_backdoor, freq_trigger, sleeper_agent, Narcissus}, with backdoor patterns applied to test samples,  assuming the attacker will not succeed when backdooring the test samples without poisoning any training data.

\textbf{Related work:} In this paper, we study DPA for practical insights into aggregation defenses against data poisoning. 
DPA is not the only aggregation defense against data poisoning but it is a good representative given its relation to others.
Related work includes: Bagging\cite{Bagging, RandomSelection}, which applies (potentially stochastic) learners to independently sampled random subsets of the training data, offering probabilistic bounds on poisoning robustness; Hash Bagging\cite{CollectiveRobust} investigates schemes to improve collective robustness of Bagging and DPA; Finite Aggregation\cite{FA}, which introduces structured overlapping among the training sets for base classifiers, providing improved robustness bounds from DPA; Run-off Election\cite{ROE}, which proposes a two-stage voting scheme that increases robustness bounds for DPA and Finite Aggregation; Temporal Aggregation\cite{temporal_robustness}, which trains base classifiers from data collected within different time frames to be certifiably robust against data poisoning attacks that are either not early enough or not long enough on the timeline. Lethal Dose Conjecture\cite{LDC} is a conjecture characterizing the limitations of poisoning robustness, implying the near optimality of DPA and other aggregation defenses when paired with the most data-efficient learners.

\section{Efficiency: Scaling Base Models for Efficient Aggregation Defenses}
\label{sec:efficiency}

Will the aggregations be computationally expensive?
Previously, aggregation defenses use a fixed model architecture with varying numbers of base models. Consequently, the computation for a single prediction scales linearly when using more base models. For example, when using $k$ partitions in DPA, the computation at inference time will be $k$ times the computation for using a single model, which is clearly not ideal for efficiency.
However, note that each base classifier learns from less training data than usual: \textbf{Why do we have to use an architecture that is as large as the one we use to learn from the entire training set?}

\subsection{Square Root Scaling of Width}
Here we present an alternative, \textbf{square root scaling of width}, which divides the width of each layer by the square root of the total number of base models, as an example implementation of this intuition. 

Why divide the width of hidden layers by the square root of the total number of base models? 
It is worth noting that this may not be the optimal scaling rule for model sizes in the context of poisoning defenses with aggregations, but it is a good starting point given its generality and simplicity.
At this moment, there are great varieties of neural network architectures, including but not limited to ResNet\cite{ResNet, wide_resnet}, VGG\cite{VGG}, DenseNet\cite{DenseNet}, EfficientNet\cite{EfficientNet}, Transformer\cite{Transformer}, Vision Transformer\cite{vit}, Swin Transformer\cite{SwinT, SwinT2}.
Despite their differences, their computations share a similar form of dependency on the width of hidden layers, i.e. a quadratic dependency.
To illustrate, we present in Table \ref{tab:resnet} the architectures of ResNet, with the width (e.g. the filter numbers for convolution layers) of layers highlighted: Since the computation for a convolution layer is proportional to the product of the number of input channels and the number of convolution filters, the bottlenecks are essentially quadratic to the width of hidden layers for all blocks except the very first convolution layer (i.e. conv1) and the linear classification head (i.e. 1000-d fc) at the end. This suggests dividing the width of hidden layers in each base model by the square root of the total number of models can be a good choice to maintain the overall efficiency of aggregations.

\newcommand{\blocka}[2]{\multirow{3}{*}{\(\left[\begin{array}{c}\text{3$\times$3, #1}\\[-.1em] \text{3$\times$3, #1} \end{array}\right]\)$\times$#2}
}
\newcommand{\blockb}[3]{\multirow{3}{*}{\(\left[\begin{array}{c}\text{1$\times$1, #2}\\[-.1em] \text{3$\times$3, #2}\\[-.1em] \text{1$\times$1, #1}\end{array}\right]\)$\times$#3}
}
\setlength{\tabcolsep}{3pt}
\begin{table*}[t]
\begin{center}
\resizebox{0.85\linewidth}{!}{
%\footnotesize
\begin{tabular}{c|c|c|c|c|c}
\hline
layer name& 18-layer & 34-layer & 50-layer & 101-layer & 152-layer \\
\hline
conv1 & \multicolumn{5}{c}{7$\times$7, \hl{64}, stride 2}\\
\hline
\multirow{4}{*}{conv2\_x}  & \multicolumn{5}{c}{3$\times$3 max pool, stride 2} \\\cline{2-6}
 & \blocka{\hl{64}}{2}  & \blocka{\hl{64}}{3} & \blockb{\hl{256}}{\hl{64}}{3} & \blockb{\hl{256}}{\hl{64}}{3} & \blockb{\hl{256}}{\hl{64}}{3}\\
  &  &  &  &  &\\
  &  &  &  &  &\\
\hline
\multirow{3}{*}{conv3\_x}  & \blocka{\hl{128}}{2}  & \blocka{\hl{128}}{4}  & \blockb{\hl{512}}{\hl{128}}{4}  & \blockb{\hl{512}}{\hl{128}}{4}  &
                              \blockb{\hl{512}}{\hl{128}}{8}\\
&  &  &  &  & \\
 &  &  &  &  & \\
\hline
\multirow{3}{*}{conv4\_x}  & \blocka{\hl{256}}{2}  & \blocka{\hl{256}}{6}  & \blockb{\hl{1024}}{\hl{256}}{6}  & \blockb{\hl{1024}}{\hl{256}}{23} & \blockb{\hl{1024}}{\hl{256}}{36}\\
  &  &  &  &  & \\
  &  &  &  &  & \\
\hline
\multirow{3}{*}{conv5\_x}   & \blocka{\hl{512}}{2}  & \blocka{\hl{512}}{3}  & \blockb{\hl{2048}}{\hl{512}}{3}  & \blockb{\hl{2048}}{\hl{512}}{3}
& \blockb{\hl{2048}}{\hl{512}}{3}\\
 &  &  &  &  & \\
  &  &  &  &  & \\
\hline
 & \multicolumn{5}{c}{average pool, 1000-d fc, softmax} \\
\hline
\end{tabular}
}
\end{center}
\caption{ResNet Architectures with width of hidden layers \hl{highlighted}. The computation overheads are quadratic to the width of hidden layers for all blocks except the first convolution layer conv1 and the linear classification head 1000-d fc at the end. This table is adapted from Table 1 by \citet{ResNet}.}
\label{tab:resnet}
\end{table*}

\subsection{Empirical Validation}
\label{sec:efficiency_empirical}
\textbf{ImageNet64$\times$64\cite{imagenet64}:} Before moving on to empirical validations, we first introduce ImageNet64$\times$64 dataset, which will be used throughout this paper. ImageNet64$\times$64 is a downsampled version of the original ImageNet\cite{ImageNet} dataset, containing all 1000 classes, 1281167 training samples, and 50000 validation samples but resized to a resolution of 64$\times$64. Thus it is much more challenging and has a much larger scale compared to ones in previous evaluations of certified poisoning defenses, such as MNIST\cite{MNIST}, CIFAR\cite{cifar}, and GTSRB\cite{GTSRB}. In addition, the reduced resolution makes it computationally more accessible for research purposes than full-resolution ImageNet.

\begin{figure}[tb!]
\begin{center}
\subfigure[Training Time v.s. Base Width of ResNet-18]{
    \includegraphics[width=0.38\linewidth]{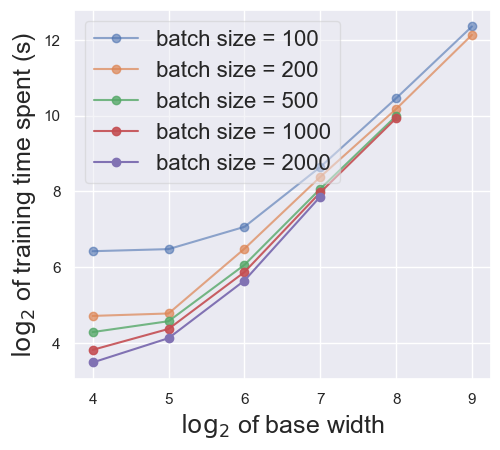}
\label{fig:resnet18_scaling_train}
}
\hfil
\subfigure[Testing Time v.s. Base Width of ResNet-18]{
    \includegraphics[width=0.4\linewidth]{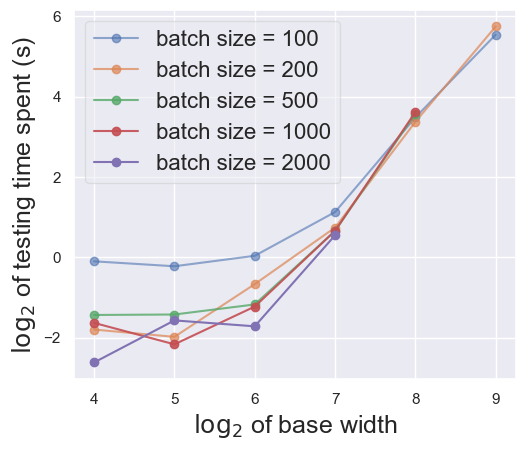}
\label{fig:resnet18_scaling_test}
}
\caption{The time spent on ResNet-18 with different widths for one epoch of training/testing on ImageNet64$\times$64 using a single RTX A4000 GPU. Both axes are plotted in logarithmic scales, where the quadratic dependency of computational overhead on the width of hidden layers are observed. A base width of $2^6=64$ corresponds to the standard ResNet-18.}
\label{fig:resnet18_scaling}
\end{center}
%\vskip -0.1in
\end{figure}

In Figure \ref{fig:resnet18_scaling}, we include the time needed for a ResNet-18 with different widths to go through one epoch of ImageNet64$\times$64 during respectively training and testing. Note that the figures are in logarithmic scales. Here the quadratic dependency of computational overhead is observed, validating our previous analysis and showing satisfying speedup when reducing model width. 

While it works as expected when the initial model is wide enough, one should also be aware of the limitation. When the width is too small, the quadratic term may no longer dominate the total computation, and the linear terms, such as the linear classification head in ResNet, will become the new bottleneck, resulting in less speedup from further reducing model width. This is also verified in Table \ref{tab:efficiency}, where we report training time and inference time for using a single ResNet-50 model with a base width of 128 and for using DPA with scaled base learners (Training details are deferred to Section \ref{sec:exp_setup} for coherence). 
However, even in Table \ref{tab:efficiency} the architecture is not wide enough to enjoy full speedup in a 1000-way classification, DPAs with scaled base learners still require less training time than the single model baseline, challenging previous beliefs on the inefficiency of aggregations.

The number of classes is also a worth noting factor: The case evaluated here consists of as many as 1000 classes, which correspond to a large constant for the linear term (i.e. the linear classification head is larger). In more common cases with fewer classes to predict, the constant is smaller and the reduced speedup corresponds to only even smaller widths.

\textbf{To summarize}, we showcase a promising direction of scaling base models for efficient aggregation defenses against data poisoning. The example scheme we presented, square root scaling of width, can greatly reduce the inference overhead of aggregation defenses, being as efficient as using a single model when the model architecture to be scaled is wide enough. Even when the model architecture is not wide enough to enjoy the full speedup of square root width scaling, DPAs with scaled base learners still require less training time than training a single model over the entire training set, yielding surprising efficiency.
In Section \ref{sec:performance} and Section \ref{sec:robustness}, we will delve into the performance and robustness aspects of aggregation defenses with scaled learners.

\begin{table}
  \caption{Training and inference time for the single model baseline (i.e. a ResNet-50 with a base width of 128, also known as Wide-ResNet-50-2) and DPA, using a single RTX A4000 GPU. The inference time is for the entire validation set of ImageNet64$\times$64, i.e. 50000 samples. Even when the architecture is not wide enough to enjoy full speedup in a 1000-way classification (details discussed in Section \ref{sec:efficiency_empirical}), DPAs with scaled base learners still require less time to train than the baseline.}
  \label{tab:efficiency}
  \centering
  \renewcommand\arraystretch{1.1}
  \begin{tabular}{ccccc}
    \toprule
         & \multirow{2}{*}{\shortstack[c]{training time\\(total)}} & \multirow{2}{*}{\shortstack[c]{training time\\(per model)}}   & \multirow{2}{*}{\shortstack[c]{ inference time \\ (batch size=128)}} &  \multirow{2}{*}{\shortstack[c]{ inference time \\ (batch size=512)}}\\
         
         &&&\\
    \hline
    baseline (single model) & 5.3 hours & 5.3 hours  & 14.8 seconds & 13.6 seconds     \\
    DPA (k=3)~~ & 3.7 hours & 1.2 hours & 28.9 seconds & 26.6 seconds\\
    DPA (k=4)~~ & 3.3 hours & 0.8 hours &  32.1 seconds & 29.9 seconds \\
    DPA (k=7)~~ & 3.0 hours & 0.4 hours &  47.2 seconds & 44.7 seconds\\
    DPA (k=9)~~ & 2.9 hours & 0.3 hours & 58.9 seconds & 56.0 seconds \\
    DPA (k=16) & 2.9 hours & 0.2 hours & 119.4 seconds & 77.9 seconds \\
    \bottomrule
  \end{tabular}
\end{table}

\section{Performance: Capability and Limitation on Preserving Utility}
\label{sec:performance}

Will the aggregations preserve utility? Intuitively, the answer should depend on both the difficulty of the objective and the quantities of available data, but what would be a proper description in this context?
In this section, we analyze the predictions of DPA and provide practical guidance in deciding the capability and limitation of aggregation defenses regarding preserving accuracy over test samples that are not under attack.

\subsection{Implementations}
\label{sec:exp_setup}
First we present the details for empirical evaluations. 

\textbf{Dataset:} We use ImageNet64$\times$64\cite{imagenet64} dataset, a downsampled version of the original ImageNet\cite{ImageNet} dataset, containing all 1000 classes, 1281167 training samples, and 50000 validation samples but resized to a resolution of 64$\times$64. The reduced resolution, while making the dataset more affordable for research purposes, increases the difficulty of classification, resulting in a task more challenging than the original ImageNet.

\textbf{Optimizer: } For the optimizer, we use SGD with one-cycle learning rate schedule\cite{onecyclelr} for 20 epochs, with the learning rate keeps increasing from 0.05 to 1.0 for the first $30\%$ of the training steps, and keeps decreasing to 0.0005 at the end of training. These hyper-parameters are adapted from the ones by \citep{onecyclelr}, which greatly reduce training time by enabling models to converge on ImageNet with 20 epochs rather than 100 epochs. 

\textbf{Model architecture: } For the single model baseline, we use ResNet-50 with a base width of 128, which is also known as Wide-ResNet-50-2\cite{wide_resnet} since the width is doubled compared to the standard ResNet-50. This choice is made through balancing efficiency and accuracy, presented in Table \ref{tab:acc_single_model}, where increasing model width from 128 to 256 results in marginally increases accuracy but more than doubled training time. For DPAs, we use the square root scaling of width proposed in Section \ref{sec:efficiency}, dividing the width of ResNet-50 by the square root of the number of partitions for base models used in aggregations. To be specific, when the number of partitions $k$ is $3,4,7,9,16$, we use ResNet-50 with a base width of respectively $74, 64, 48, 42, 32$ (approximately $128/\sqrt{k}$) to learn from each partition.

\begin{table}
  \caption{Training time and accuracy of ResNet-50 with different widths when trained with the entire training set (i.e. the single model baselines). `acc@n' denotes top-n accuracy.}
  \label{tab:acc_single_model}
  \centering
  \renewcommand\arraystretch{1.2}
  \resizebox{0.95\linewidth}{!}{
  \begin{tabular}{lcccccccc}
    \hline
         & training time & acc@1 & acc@3 & acc@5 & acc@10 & acc@20 & acc@50 & acc@100\\
    \hline
    ResNet-50 (width=64) & ~~3.4 hours & 50.72\% & 68.72\% & 75.00\% & 82.25\% & 88.21\% & 93.67\% & 96.64\%\\
    ResNet-50 (width=128) & ~~5.3 hours & 52.53\% &
    70.27\% & 76.57\% & 83.71\% & 89.14\% & 94.35\% & 97.03\%\\
    ResNet-50 (width=256) & 12.8 hours & 53.52\% & 71.42\% & 77.32\% & 84.17\% & 89.39\% & 94.38\% & 97.08\%\\
    \hline
  \end{tabular}
  }
\end{table}

\textbf{Hash function for DPA: }
Many aggregation defenses against poisoning attacks define the subsets of training data for base models in a deterministic way with the help of a hash function.  \citet{DPA} uses the sum of pixel values for hashing purposes in DPA and this protocol is followed by many\cite{FA, ROE, LDC}. While using the sum of pixel values is reasonable for research, informally we find that it leads to certain confusions in the research community, where it is improperly assumed that attackers can easily choose to poison training samples for a specific base model in aggregations by manipulating hash values (e.g. sum of pixels). Here we present a very simple and general construction of hash functions for vectors, with bounded collision rates.

\begin{definition}[Weighted Sum for Hashing]
Given a field $F$ and a $n$-dimensional weight vector $W\in F^n$, the hash value for any $X\in F^n$ is defined as $h_{W}(X) = W^T X = \sum_{i=1}^n W_i X_i$.
\end{definition}
This is simply the weighted version for `sum of pixels'. However, it offers rigorous guarantees for collision rates with no assumption on data distributions, through Schwartz-Zippel Lemma.

\begin{theorem}[Schwartz-Zippel Lemma\cite{demillo1978probabilistic, zippel1979probabilistic, schwartz1980fast}]
Let $P \in F[x_1, x_2, \dots, x_n]$ be a non-zero polynomial of degree $d$ over a field $F$. Let $S$ be a finite subset of $F$ and $w_1, w_2, \dots, w_n$ be selected uniformly and independently at random from $S$, then $\Pr[P(w_1, w_2, \dots, w_n) = 0] \leq d / |S|$.
\end{theorem}
For implementation, one can first pick a large prime $p$ and use $F_p$ (i.e. integers modulo p) as the underlying field $F$, and select dimensions of the weight vector $W$ uniformly and independently from $S = F_p$ (i.e. from $0, 1, \dots, p-1$). Then for any $X, X'\in F^n$ such that $X\neq X'$, $h_W(X) - h_W(X')$ can be viewed as a non-zero polynomial of $W_1, W_2, \dots, W_n$ with a degree of 1, and therefore $\Pr[h_W(X) = h_W(x')]\leq 1 / |S| = 1 / p$. In our experiments, we use the first prime $p$ larger than $2^{48}$ and use two independently sampled weight vectors for hashing purposes, resulting in a probability of no more than $\binom{N}{2} \cdot 1/p^2\approx 10^{-17}$ for any collision of hash values to happen within the training set, where $N=1281167$ is the total number of samples. In addition, we empirically verify that there is no hash collision at all in our experiments.

Besides the increased difficulty of manipulation, another benefit of the bounded collision is to enable more efficient sorting of training data. Depending on the underlying assumptions for the data collection process, the order of samples in the training set may or may not be preserved after poisoning. 
Consequently, many aggregation defenses \cite{DPA, FA, CollectiveRobust, ROE, LDC} require the samples to be sorted deterministically for certified robustness when the orders may not be preserved after poisoning. Previously, samples are sorted lexicographically according to input dimensions, leading to a worst-case complexity of $O(DN\log N)$, with $D$ being sample dimensions and $N$ being the number of samples. When there is no collision for hash values, this is reduced to $O(N(\log N + D))$, by simply sorting samples according to hash values.

\subsection{Practical Guidance for Performance with the Data-to-Complexity Ratio}
\label{sec:performance_analysis}
\begin{table}
  \caption{The accuracy comparisons of baseline and DPAs. `acc@n' denotes top-n accuracy. \hl{Differences} from the single model baseline are \hl{highlighted}. Viewing top-n accuracy with different n as tasks with objectives of different difficulties, one can see the capability and limitation of aggregation defenses indeed depend on difficulties of the objective, with the accuracy drops being larger with difficult objectives (e.g. fine-grained ones like acc@1) and smaller with easier objectives (e.g. coarse-grained ones like acc@100). Details discussed in Section \ref{sec:performance_analysis}. }
  \label{tab:acc_DPA}
  \centering
  \renewcommand\arraystretch{1.2}
  \resizebox{0.95\linewidth}{!}{
  \begin{tabular}{c|ccccccc}
    \hline
         & acc@1 & acc@3 & acc@5 & acc@10 & acc@20 & acc@50 & acc@100\\
    \hline
    baseline (single model) & 52.53\% &
    70.27\% & 76.57\% & 83.71\% & 89.14\% & 94.35\% & 97.03\%\\
    \hline
    \multirow{2}{*}{DPA (k=3)~~} & 35.92\% & 53.08\% & 60.40\% & 69.78\% & 78.12\% & 87.13\% & 92.70\%\\
    & \hl{~16.61\%~} & \hl{~17.19\%~} & \hl{~16.17\%~}& \hl{~13.93\%~}& \hl{~11.02\%~}& \hl{~~~7.22\%~}& \hl{~~~4.33\%~}\\
    \hline
    \multirow{2}{*}{DPA (k=4)~~} & 32.06\% & 48.89\% & 56.45\% & 66.79\% & 75.95\% & 86.17\% & 91.95\%\\
    & \hl{~20.47\%~} & \hl{~21.38\%~} & \hl{~20.12\%~}& \hl{~16.92\%~}& \hl{~13.19\%~}& \hl{~~~8.18\%~}& \hl{~~~5.08\%~}\\
    \hline
    \multirow{2}{*}{DPA (k=7)~~} & 25.96\% & 41.64\% & 49.20\% & 59.61\% & 69.35\% & 81.10\% & 88.72\%\\
    & \hl{~26.57\%~} & \hl{~28.63\%~} & \hl{~27.37\%~}& \hl{~24.10\%~}& \hl{~19.79\%~}& \hl{~13.25\%~}& \hl{~~~8.31\%~}\\
    \hline
    \multirow{2}{*}{DPA (k=9)~~} & 23.57\% & 38.36\% & 45.62\% & 55.66\% & 66.21\% & 78.97\% & 86.85\%\\
    & \hl{~28.96\%~} & \hl{~31.91\%~} & \hl{~30.95\%~}& \hl{~28.05\%~}& \hl{~22.93\%~}& \hl{~15.38\%~}& \hl{~10.18\%~}\\
    \hline
    \multirow{2}{*}{DPA (k=16)} & 18.29\% & 31.28\% & 38.11\% & 48.42\% & 59.18\% & 73.46\% & 83.17\%\\
    & \hl{~34.24\%~} & \hl{~38.99\%~} & \hl{~38.46\%~}& \hl{~35.29\%~}& \hl{~29.96\%~}& \hl{~20.89\%~}& \hl{~13.86\%~}\\
    \hline
  \end{tabular}
  }
\end{table}

\textbf{How good are aggregation defenses in preserving utility for tasks with different difficulties?}

We present accuracy comparisons of DPAs with the single model baseline in Table \ref{tab:acc_DPA}, including top-n accuracy with n$\in\{1,3,5,10,20,50,100\}$. Viewing top-n accuracy with different n as tasks with objectives of different difficulties, Table \ref{tab:acc_DPA} provides an overview of the limitation of aggregation defenses in preserving accuracy. Notably, for n>1, acc@n for DPAs are computed by counting the top-n predictions from each base classifier during aggregations, to be consistent with viewing the objective as maximizing the probability of one hit within n guesses (i.e. top-n accuracy). 

Here one can see both the limitation and the potential. On the one hand, the accuracy drops can be unacceptable (e.g. 16.61\%$\sim$34.24\% drops of acc@1 from a baseline of 52.53\%) when targeting very fine-grained classifications, suggesting clear limitations for applicability (given existing learners). On the other hand, the accuracy drops are fairly marginal (e.g. 4.33\%$\sim$13.86\% drops of acc@100 from a baseline of 97.03\%) for coarse-grained cases, suggesting the potential of aggregation defenses for large-scale datasets with millions of samples.

\begin{table}[tbp!]
  \caption{The accuracy comparisons of DPAs and base classifiers trained using different portions of the training data. The \hl{differences} of accuracy are \hl{highlighted}. While DPA improves accuracy over a single model trained from the corresponding portion of data, the improvements are somewhat limited for cases where DPA accuracy is high, suggesting \textbf{data-to-complexity ratio}, i.e. the ratio between the data set size and the sample complexity, as a practical estimation of the maximum number of partitions k for DPA to preserve accuracy. Details discussed in Section \ref{sec:performance_analysis}. }
  \label{tab:acc_base_learner}
  \centering
  \renewcommand\arraystretch{1.2}
  \resizebox{0.95\linewidth}{!}{
  \begin{tabular}{c|ccccccc}
    \hline
         & acc@1 & acc@3 & acc@5 & acc@10 & acc@20 & acc@50 & acc@100\\
    \hline
    \multirow{1}{*}{DPA (k=3)~~} & 35.92\% & 53.08\% & 60.40\% & 69.78\% & 78.12\% & 87.13\% & 92.70\%\\
    single model (1/3 data)~~ & 32.88\% & 49.02\% & 56.28\% & 65.82\% & 74.70\% & 84.76\% & 90.92\%\\
    & \hl{~~~3.04\%~} & \hl{~~~4.06\%~} & \hl{~~~4.12\%~}& \hl{~~~3.96\%~}& \hl{~~~3.42\%~}& \hl{~~~2.37\%~}& \hl{~~~1.78\%~}\\
    \hline
    \multirow{1}{*}{DPA (k=4)~~} & 32.06\% & 48.89\% & 56.45\% & 66.79\% & 75.95\% & 86.17\% & 91.95\%\\
    single model (1/4 data)~~ & 24.02\% & 38.41\% & 45.44\% & 55.25\% & 65.19\% & 77.58\% & 85.71\%\\
    & \hl{~~~8.04\%~} & \hl{~10.48\%~} & \hl{~11.01\%~}& \hl{~11.54\%~}& \hl{~10.76\%~}& \hl{~~~8.59\%~}& \hl{~~~6.24\%~}\\
    \hline
    \multirow{1}{*}{DPA (k=7)~~} & 25.96\% & 41.64\% & 49.20\% & 59.61\% & 69.35\% & 81.10\% & 88.72\%\\
    single model (1/7 data)~~ & 18.56\% & 30.90\% & 37.64\% & 47.25\% & 57.48\% & 71.38\% & 81.20\%\\
    & \hl{~~~7.40\%~} & \hl{~10.74\%~} & \hl{~11.56\%~}& \hl{~12.36\%~}& \hl{~11.87\%~}& \hl{~~~9.72\%~}& \hl{~~~7.52\%~}\\
    \hline
    \multirow{1}{*}{DPA (k=9)~~} & 23.57\% & 38.36\% & 45.62\% & 55.66\% & 66.21\% & 78.97\% & 86.85\%\\
    single model (1/9 data)~~ & 15.40\% & 26.16\% & 32.37\% & 41.75\% & 52.07\% & 66.48\% & 77.33\%\\
    & \hl{~~8.17\%~} & \hl{~12.20\%~} & \hl{~13.23\%~}& \hl{~13.91\%~}& \hl{~14.14\%~}& \hl{~12.49\%~}& \hl{~~~9.52\%~}\\
    \hline
    \multirow{1}{*}{DPA (k=16)} & 18.29\% & 31.28\% & 38.11\% & 48.42\% & 59.18\% & 73.46\% & 83.17\%\\
    single model (1/16 data) & 10.93\% & 19.71\% & 25.14\% & 33.89\% & 44.23\% & 59.98\% & 72.27\%\\
    & \hl{~~~7.36\%~} & \hl{~11.57\%~} & \hl{~12.97\%~}& \hl{~14.53\%~}& \hl{~14.95\%~}& \hl{~13.48\%~}& \hl{~10.90\%~}\\
    \hline
  \end{tabular}
  }
\end{table}

\textbf{How many partitions/base models are allowed when preserving accuracy?}

Notably, DPA with only $k=1$ partition is essentially the single model baseline, and increasing the number of partitions $k$ results in decreased accuracy, as in Table \ref{tab:acc_DPA}. Thus the core for understanding the performance of DPA is to understand how many partitions are allowed while preserving accuracy. 

Towards this end, we present in Table \ref{tab:acc_base_learner} the comparisons of DPAs with base classifiers trained from different portions of the training set, where we again, use top-n accuracy with different n as tasks with objectives of varying difficulty. While DPA improves accuracy over a single model trained from the corresponding portion of data, the improvements are somewhat limited, specifically for cases where DPA accuracy is high. This observation suggests the practical necessity for base learners to have decent accuracy for the aggregations to be accurate, as opposed to the theoretical possibility for aggregations to be accurate with lowly performed but diverse base models.

Considering the accuracy of base learners as pessimistic but acceptable approximations of the performance of aggregations, the maximum number of partitions $k$ allowed while preserving accuracy will be approximated by the data-to-complexity ratio, which is the ratio between the dataset size and sample complexity of the objective (i.e. the number of samples needed to train classifiers with accuracy approaching the target accuracy). This provides an empirical approach to deciding the number of base models through training base models with small fractions of data instead of training many aggregations: Per Table \ref{tab:efficiency}, training one base model takes only a very small amount of time compared to the training of entire aggregations and to the training of the single model baseline.

\section{Robustness: How Aggregations Boost Empirical Poisoning Robustness}
\label{sec:robustness}

How can the aggregations boost empirical robustness?
Previously, the evaluations of \citet{baracaldo2022benchmarking} suggest that aggregation defenses, specifically DPA\cite{DPA} and Finite Aggregation\cite{FA}, can be fairly effective compared to others in defending various empirical attacks of data poisoning, including both dirty-label attacks\cite{badnets,Trojan} and clean-label attacks\cite{CLBD, Brew}. However, the key mechanism behind the empirical robustness of aggregation defenses is different from the motivation for its certified robustness and is never revealed previously. In this section, we will present the poisoning overfitting phenomenon and its role to the empirical robustness of aggregations..

\subsection{Evaluation Settings}

To assess the empirical robustness of aggregation defenses to poisoning attacks, we utilize a simple form of backdoor attacks, with the backdoor trigger being simply a black square patch in the center of the target images. Two different patch sizes are evaluated, with the settings with 16$\times$16 patches representing attacks emphasizing effectiveness and the settings with $1\times1$ patches representing attacks emphasizing subtlety. For a given attack budget (i.e. a given percentage of the training set to be poisoned), a random subset of the training data with the corresponding size will be patched, with their labels changed into class 0, the target class of the attack. The attack success rate is measured by patching the entire validation set with the backdoor pattern and computing the fraction of predictions matching the target class (i.e. class 0).

Such simple attacks are not considered strong in general, given that even small patches can be somewhat visible and the labels of the poisoned training samples will not match the semantics. However, they are strong attacks when evaluating defenses without (potentially implicit) data filtering or outlier detection, including aggregation defenses (with no other defense incorporated).
Thus we choose to use the simple patch backdoors to assess the empirical robustness of aggregation defenses.

\newcommand{\nhl}[1]{\textcolor{BrickRed}{#1}}
\begin{table}
  \caption{The attack success rates of DPAs and the single model baseline, when being attacked by backdoor attacks with different levels of visibility and different attack budgets (\%poison denotes the percentage of poisoned samples in the training set; \#poison denotes the number of poisoned samples). The amount of \textbf{reduced attack success rates} compared to the baseline are \textbf{highlighted}, with \textbf{improved robustness} in \hl{blue} and \textbf{degraded robustness} in \nhl{red}. DPAs improve empirical robustness to poisoning attacks over the single model baseline and increasing the number of base models $k$ leads to further reduced attack success rates.}
  \label{tab:robustness}
  \centering
  \renewcommand\arraystretch{1.2}
  \resizebox{0.97\linewidth}{!}{
  \begin{tabular}{c|c|cccccc}
    \hline
    \multirow{3}{*}{attack type} & \multirow{3}{*}{defense type} & \multicolumn{6}{c}{attack success rate ($\downarrow)$}\\
    \cline{3-8}
    & & 
    \multirow{2}{*}{\shortstack[c]{\%poison = 0.0025 \\ \#poison = 32}}&
    \multirow{2}{*}{\shortstack[c]{\%poison = 0.005 \\ \#poison = 64}}&
    \multirow{2}{*}{\shortstack[c]{\%poison = 0.01 \\ \#poison = 128}}&
    \multirow{2}{*}{\shortstack[c]{\%poison = 0.03 \\ \#poison = 384}}&
    \multirow{2}{*}{\shortstack[c]{\%poison = 0.1 \\ \#poison = 1281}}&
    \multirow{2}{*}{\shortstack[c]{\%poison = 0.3 \\ \#poison = 3843}}
    \\
    & & \\
    \hline \hline
    \multirow{12}{*}{\shortstack[c]{16$\times$16\\ backdoor}} & \multirow{2}{*}{\shortstack[c]{baseline \\(single model)}} & \multirow{2}{*}{70.27\%} &
    \multirow{2}{*}{85.22\%} & \multirow{2}{*}{95.04\%} & \multirow{2}{*}{98.27\%} & \multirow{2}{*}{99.49\%} & \multirow{2}{*}{99.76\%}\\
    & & \\
    \cline{2-8}
    & \multirow{2}{*}{DPA (k=3)~~} & 54.20\% & 82.23\% & 90.78\% & 97.56\% & 99.06\% & 99.62\%\\
    & & \hl{~16.07\%~} & \hl{~~~2.99\%~} & \hl{~~~4.26\%~}& \hl{~~~0.71\%~}& \hl{~~~0.43\%~}& \hl{~~~0.14\%~}\\
    \cline{2-8}
    & \multirow{2}{*}{DPA (k=4)~~} & 56.86\% & 78.34\% & 92.21\% & 96.45\% & 99.08\% & 99.66\%\\
    & & \hl{~13.41\%~} & \hl{~~~6.88\%~} & \hl{~~~2.83\%~}& \hl{~~~1.82\%~}& \hl{~~~0.42\%~}& \hl{~~~0.10\%~}\\
    \cline{2-8}
    & \multirow{2}{*}{DPA (k=7)~~} & 47.36\% & 75.21\% & 89.70\% & 97.34\% & 99.03\% & 99.61\%\\
    & & \hl{~22.91\%~} & \hl{~10.01\%~} & \hl{~~~5.34\%~}& \hl{~~~0.93\%~}& \hl{~~~0.46\%~}& \hl{~~~0.15\%~}\\
    \cline{2-8}
    & \multirow{2}{*}{DPA (k=9)~~} & 50.64\% & 64.57\% & 88.70\% & 97.24\% & 98.92\% & 99.61\% \\
    & & \hl{~19.63\%~} & \hl{~20.65\%~} & \hl{~~~6.34\%~}& \hl{~~~1.03\%~}& \hl{~~~0.57\%~}& \hl{~~~0.15\%~}\\
    \cline{2-8}
    & \multirow{2}{*}{DPA (k=16)} & 22.51\% & 56.71\% & 83.50\% & 97.25\% & 99.00\% & 99.61\%\\
    & & \hl{~47.76\%~} & \hl{~28.51\%~} & \hl{~11.54\%~}& \hl{~~~1.02\%~}& \hl{~~~0.49\%~}& \hl{~~~0.15\%~}\\
    \hline \hline
    
    \multirow{12}{*}{\shortstack[c]{1$\times$1\\ backdoor}} & \multirow{2}{*}{\shortstack[c]{baseline \\(single model)}} & \multirow{2}{*}{~~0.12\%} &
    \multirow{2}{*}{~~0.13\%} & \multirow{2}{*}{~~0.12\%} & \multirow{2}{*}{~~0.52\%} & \multirow{2}{*}{67.36\%} & \multirow{2}{*}{80.89\%}\\
    & & \\
    \cline{2-8}
    & \multirow{2}{*}{DPA (k=3)~~} & ~~0.20\% & ~~0.20\% & ~~0.25\% & ~~0.39\% & 15.45\% & 75.04\%\\
    & & \nhl{~~-0.08\%~} & \nhl{~~-0.07\%~} & \nhl{~~-0.13\%~}& \hl{~~~0.13\%~}& \hl{~51.91\%~}& \hl{~~~5.85\%~}\\
    \cline{2-8}
    & \multirow{2}{*}{DPA (k=4)~~} & ~~0.21\% & ~~0.25\% & ~~0.25\% & ~~0.37\% & ~~4.48\% & ~66.90\%\\
    & & \nhl{~~-0.09\%~} & \nhl{~~-0.12\%~} & \nhl{~~-0.13\%~}& \hl{~~~0.15\%~}& \hl{~62.88\%~}& \hl{~13.99\%~}\\
    \cline{2-8}
    & \multirow{2}{*}{DPA (k=7)~~} & ~~0.24\% & ~~0.27\% & ~~0.27\% & ~~0.39\% & ~~1.91\% & 29.08\%\\
    & & \nhl{~~-0.12\%~} & \nhl{~~-0.14\%~} & \nhl{~~-0.15\%~}& \hl{~~~0.13\%~}& \hl{~65.45\%~}& \hl{~51.81\%~}\\
    \cline{2-8}
    & \multirow{2}{*}{DPA (k=9)~~} & ~~0.26\% & ~~0.25\% & ~~0.27\% & ~~0.27\% & ~~0.96\% & 16.93\% \\
    & & \nhl{~~-0.14\%~} & \nhl{~~-0.12\%~} & \nhl{~~-0.15\%~}& \hl{~~~0.25\%~}& \hl{~66.40\%~}& \hl{~63.96\%~}\\
    \cline{2-8}
    & \multirow{2}{*}{DPA (k=16)} & ~~0.24\% & ~~0.22\% & ~~0.23\% & ~~0.20\% & ~~0.35\% & ~~2.13\%\\
    & & \nhl{~~-0.12\%~} & \nhl{~~-0.09\%~} & \nhl{~~-0.11\%~}& \hl{~~~0.32\%~}& \hl{~67.01\%~}& \hl{~78.76\%~}\\
    \hline\hline
  \end{tabular}
  }
\end{table}

\subsection{Key Mechanism: The Hidden Assumption for Poisoning Overfitting}
In Table \ref{tab:robustness}, we report the attack success rates for DPA when being attacked by backdoor attacks with different levels of visibility and different attack budgets.

Firstly, one can verify that aggregation defenses like DPA are empirically robust to poisoning attacks with budgets much larger than their certifications, which is consistent with previous empirical evaluations\cite{baracaldo2022benchmarking}: Theorem \ref{th:DPA} suggests DPA can be certifiably robust with the existence of at most $k/2 = O(k)$ poisoned samples, while in Table \ref{tab:robustness}, DPA can effectively reduce attack success rates when there are 32$\sim$64 samples poisoned by the effectiveness-emphasizing attacks (with backdoor patches of size 16$\times$16) or 1281$\sim$3843 samples poisoned by the subtlety-emphasizing attacks (with backdoor patches of size 1$\times$1), much larger than the corresponding values of $k$. It is understood how this can be the case, since empirically it may take more than one poisoned sample to change the prediction of a single base model, unlike the assumption used in computing bounds of certified poisoning robustness from aggregation defenses.

However, how aggregation defenses can be empirically more robust to poisoning attacks than the single model baseline was never explained previously. Here we present the poisoning overfitting phenomenon, which is the key for aggregation defenses to boost poisoning robustness empirically when already exceeding the range of their certified robustness.
\begin{definition}[Poisoning Overfitting]
Poisoning overfitting refers to the phenomenon where the attack success rate decreases when scaling down the size of the training set while keeping the percentage of poisoned samples unchanged.
\end{definition}
Notably, this phenomenon depends on both the learning algorithm and the schemes of poisoning attacks. Informally, the poisoning overfitting phenomenon suggests certain generalization requirements for the attack to succeed. Thus when the absolute number of poisoned samples is reduced, even though the relative amount of poisoned samples (i.e. the percentage of poisoned samples) remains unchanged, the learner can still overfit the small set of poisoned samples, resulting in poor generalization at test time and therefore reduced attack success rates.

Another way to understand poisoning overfitting is to consider cases where it is not true. If the attack success rates remain when keeping the percentage of poisoned samples, DPA and other aggregation defenses will not boost empirical robustness against data poisoning: Assuming samples are divided almost uniformly into different partitions, the training set of every base learner will on average still contain the same percentage of poisoned samples and therefore no extra robustness should be expected. This is why poisoning overfitting should be considered as an assumption about both the learner and the attacks when using aggregation defenses to improve robustness against data poisoning attacks. The discovery of poisoning overfitting opens up some new potential directions for the study of poisoning attacks and defenses, including but not limited to designing advanced attacks that avoid poisoning overfitting to prevent aggregation defenses from enhancing robustness, contrasting or connecting poisoning overfitting with the assumptions of other poisoning defenses for combined defenses robust to a broader range of attacks, and comparing poisoning overfitting with different learning algorithms to develop learners better suited for defending against poisoning attacks.
\section{Conclusion}
\label{sec:conclusion}

In this work, we focus on practical aspects of aggregation defenses against data poisoning. To improve the efficiency of aggregation defenses, we demonstrate square root scaling of width as a simple yet practical approach for scaling base models. We propose and empirically support the data-to-complexity ratio as a practical estimation of the allowed number of base models while preserving overall performance, providing guidance for determining the capability and limitation with minimal effort. Last but not least, we assess the empirical robustness of aggregation defenses and point out the phenomenon of poisoning overfitting, which is the previously unknown underlying mechanism allowing aggregation defenses to boost empirical robustness against poisoning attacks while already exceeding the scope of their certified robustness. The discovery of poisoning overfitting phenomenon naturally opens up new directions for future research, including designing attacks avoiding the phenomenon to target aggregation defenses, contrasting the phenomenon to other defense assumptions, and understanding the role of learners in the phenomenon for potential improvements.

\section*{Acknowledgements}
This project was supported in part by Meta grant 23010098, NSF CAREER AWARD 1942230, HR001119S0026 (GARD), ONR YIP award N00014-22-1-2271, Army Grant No. W911NF2120076, NIST 60NANB20D134, the NSF award CCF2212458, a capital one grant and an Amazon Research Award.

%%%%%%%%%%%%%%%%%%%%%%%%%%%%%%%%%%%%%%%%%%%%%%%%%%%%%%%%%%%%

\bibliographystyle{plainnat}
\bibliography{ref}
\nocite{*}

\end{document}